\documentclass{article}

\usepackage{subcaption}
\usepackage[table]{xcolor}
\usepackage{hyperref}
\usepackage{enumitem}
\usepackage{arxiv}

\usepackage[utf8]{inputenc} 
\usepackage[T1]{fontenc}    
\usepackage{hyperref}       
\usepackage{url}            
\usepackage{booktabs}       
\usepackage{amsfonts}       
\usepackage{nicefrac}       
\usepackage{microtype}      
\usepackage{lipsum}
\usepackage{graphicx}

\title{A Hybrid Framework for Matching\\Printing Design Files to Product Photos}

\author{
\textbf{Alper KAPLAN$^{1}$, Erdem AKAGÜNDÜZ$^{2}$}\\
$^{1}$Cognitive Science Program, Graduate School of Social Sciences, Yeditepe University, Istanbul, Turkey\\
$^{2}$Department of Electrical and Electronics Engineering, Çankaya University, Ankara, Turkey
\\
}

\begin{document}
\maketitle

\begin{abstract} We propose a real-time image matching framework, which is hybrid in the sense that it uses both hand-crafted features and deep features obtained from a well-tuned deep convolutional network. The matching problem, which we concentrate on, is specific to a certain application, that is, printing design to product photo matching. Printing designs are any kind of template image files, created using a design tool, thus are perfect image signals. However, photographs of a printed product suffer many unwanted effects, such as uncontrolled shooting angle, uncontrolled illumination, occlusions, printing deficiencies in color, camera noise, optic blur, et cetera. For this purpose, we create an image set that includes printing design and corresponding product photo pairs with collaboration of an actual printing facility. Using this image set, we benchmark various hand-crafted and deep features for matching performance and propose a framework in which deep learning is utilized with highest contribution, but without disabling real-time operation using an ordinary desktop computer.  

\end{abstract}

\section{Introduction}
\label{Int}
Image matching is a broad title that covers or partially relates to various topics among a number of different computer vision problems, namely image-based localization, multi-view 3D reconstruction, structure-from-motion, image retrieval, tracking, just to name a few. This title may refer to finding a transformed version of an image \cite{Dharani13}, or may allude to a different version of the problem, such as finding an image with a similar semantic context \cite{Liu2007}. Regardless of the problem definition, image matching boils down to a simple statement: {finding a similarity model between (at least) two images, which would satisfy the pairings for a given image set}.

In recent years, a number of algorithms have been proposed under this title, which can be mainly split into two principle categories. The first category consists of approaches that utilize hand-crafted representations. Among these methods, the bag-of-words (BoW) algorithm \cite{Sivic2003} proved to be very successful, irrespective of the type of the hand-crafted feature used, and is still the state-of-the-art approach due to its flexibility, compactness and speed. 

However, as a part of the growing wave of interest on deep-learning-based methods, a second category of approaches recently focus on image matching using convolutional neural networks (CNN) \cite{Wang15}. The strength of these methods comes from the abstract features that emerge at the deeper layers of CNNs \cite{Yosinski2015}. The earlier approaches of this category \cite{Simonyan15,Szegedy15,Krizhevsky12,Sermanet13} performs particularly good at problems like image category classification, object detection and/or localization, mainly because of their capability to convolve abstract features into image categories or object definitions. There are also attempts with promising results, which aim at transferring pre-trained and well-tuned CNNs into image retrieval frameworks \cite{Babenko14,Chandrasekhar16}. Nonetheless, these network structures are not well-suited to match a given image to its pair, since they use fully connected layers that lead to a classification layer (such as \emph{soft-max}). This final layer is used to classify the extracted abstract features into object categories. Therefore, their structure is not designed with the purpose of finding pairs. 

Very recently a new CNN structure, namely the Siamese network (SN), has been proposed specifically for the problem of image matching \cite{Melekhov16}. SNs utilize labeled training image pairs to learn an image-level feature representation so that similar images are mapped close to each other in the feature space, where dissimilar image pairs are mapped far from each other. This approach has also been successfully applied to similar problems that require an image-to-image matching, such as face recognition \cite{Taigman14} or aerial-to-ground image matching \cite{Lin15}. SNs improve matching performance dramatically, however they come up with two main drawbacks. Firstly, they necessitate the creation of a large-scale image set, because in order to span the entire space of possible transformations from the original image to the image to be matched, a massive number of image pairs are required. Such an image set collection and annotation effort is extremely expensive and usually, industrially impracticable. Secondly, these networks make a separate full forward deep CNN run for each candidate image in the image set, thus are slow even with dedicated hardware, such as a GPU.

Deep learning is a powerful tool. In less than a decade, nearly all vision problems shifted to CNN domain.  Nevertheless, it still comes with a price. As the layers of a CNN get deeper, the hardware requirements for real-time operation become more and more expensive. We still don't have a mobile solution, which may replace the high-cost and power-hungry GPUs that allow real-time deep learning operations. And when it comes to the problem of image matching, our best solution yet, namely the Siamese networks architecture, require massive training sets and forward-run for all possible candidate images. In conclusion we still need ingenious solutions for real-time, operation-specific and affordable image matching frameworks.

\subsection{Problem Definition and the Proposed Solution}
In this paper we study a particular version of the image matching problem, in which we match printing design files to product photos. Printing designs are any kind of template image files, created using a design tool and used as templates for printing a flyer, banner, poster, etc. These files are computer generated, thus they possess no signal-based deficiencies like noise or optic blur. Most of them are in vector format, hence, are resolution-free. 

On the other hand, photographs of a printed product suffer many unwanted effects, such as uncontrolled shooting angle, uncontrolled illumination, occlusions, printing deficiencies in color, camera noise, optic blur, etc. Matching them to their original design files requires learning the unknown transformation that the photographing action creates. This transformation is not deterministic by nature and is affected by predominant uncontrolled factors, such as the photographer, the camera or the background.

\begin{figure*}[t]
\centering
\includegraphics*[clip=false,width=1\textwidth]{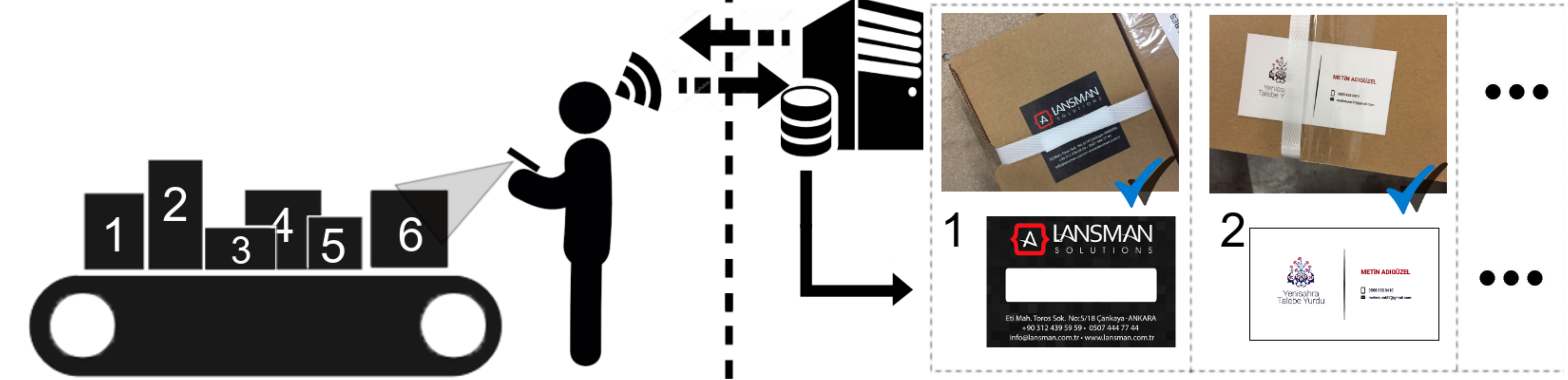}
\caption{Pictorial representation of our operational problem definition.}
 \label{operation}
\end{figure*}

A pictorial representation of our problem definition and system framework is depicted in Figure \ref{operation}. In a sample scenario of our problem definition, an operator (or an automatic visualization system) shoots the photographs of some printed products by using a computation-limited device (such as a mobile phone, etc). Then this device sends the product photo to a server machine, in which the photo is matched to its design file pair, in real-time.

In order to solve this problem, we propose a real-time image matching framework, which is hybrid in the sense that it uses both hand-crafted features and deep features obtained from a well-tuned very deep CNN \cite{Simonyan15}. The hand-crafted or deep features are extracted only from a region that is designated by a fine-tuned deep CNN. In our framework, this feature region segmentation operation is the only ``deep'' operation that is applied on the product photo, thus we avoid running a deep CNN for each possible pair in the image set, as it is done for Siamese networks. This also prevents us from using expensive deep learning hardware (a GPU), but still permits us to provide real-time operation. By using the hand-crafted or deep features extracted from the deep segmented region, a BoW framework is utilized in order to find the correct image pair. 

The rest of the paper is organized as follows: Section \ref{imagesetsection} provides the details of the image set that includes printing design files and corresponding product photo pairs. Section \ref{deepseg} explains the deep learning experiments, which aim at solving feature region segmentation problem. Section \ref{ImgMatFWork} represents the BoW framework, in which different hand-crafted and/or deep feature extraction, and product segmentation methods are benchmarked for optimal performance. Section \ref{ExpRes} presents the experimental results, whereas the final section concludes the paper and gives directions for future work. 

\section{Design File - Product Photo Pairs Image Set}
\label{imagesetsection}
The existing image sets \cite{Deng17,Datta2005,Clough2005,Schaefer17} prepared for matching or retrieval problems in the literature are very diverse in category. They deal with different problems such as retrieving RAW images, medical images, outdoor images, or even satellite images. Consequently, for each image set the problem definition is different. That's why, in order to provide a solution for our specific problem definition, we need to create a specific image set that includes printing design files and corresponding product photos.

\begin{figure*}[t]
\centering
\includegraphics*[clip=false,width=1\textwidth]{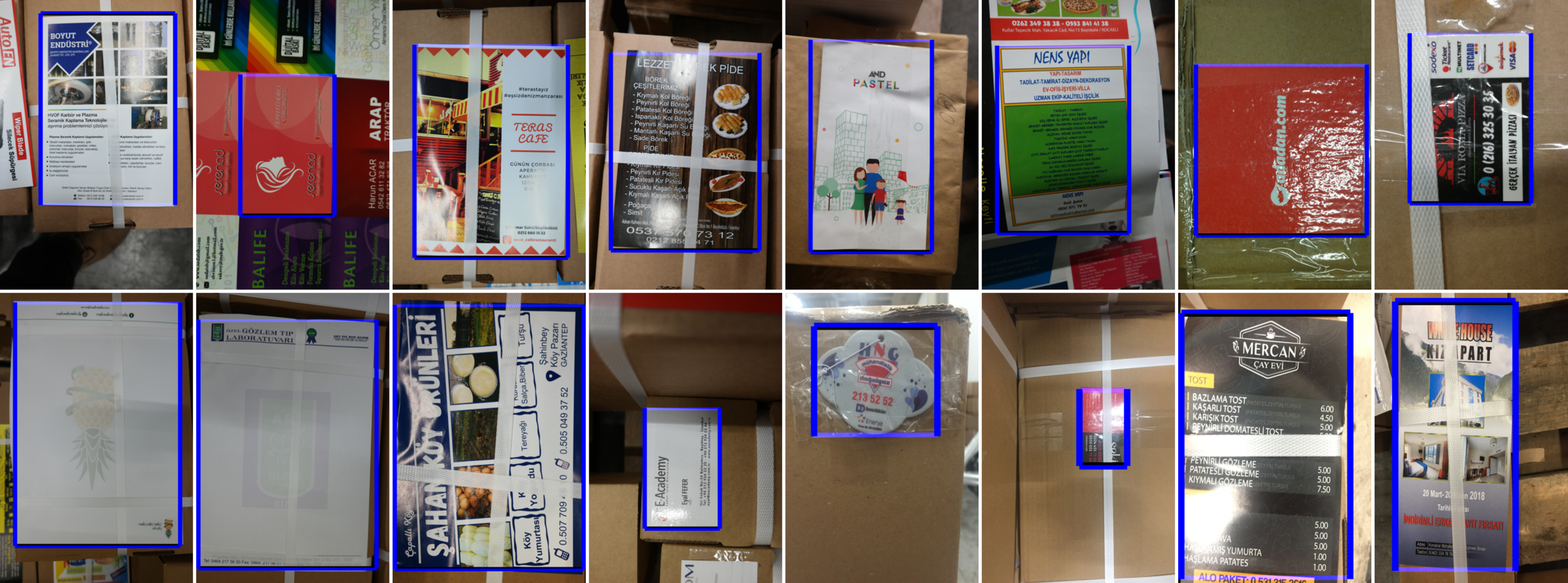}
\caption{Sample product photos from the image set.}
 \label{dataset}
\end{figure*}

As a consequence, an image set creation effort was carried out. To this end, an operator took the photographs of 2000 products at the production line. Product photos are images of a sample product (e.g. a flyer) usually stitched over a cargo box, which carries the other printed samples (Figure \ref{dataset}). The idea is to recognize this product (i.e. its ID) by matching the image of the sample on the cargo box with the design file at the server.

For some of the products, there exists more than a single design file. A good example is a business card (Figure \ref{operation}), which usually has information on both sides, and thus has two separate design files. In these cases, the problem definition is to match the product photo (which could be any face of the card) to one of the design files in the image set. Consequently, for the photographed 2000 product samples, 3458 design files were added to the image set. The operators were \emph{advised} to shoot the product with a perpendicular angle so that the product (usually, but not necessarily rectangular in shape) would fit the image with uniform margins. However this weak protocol was not successfully applied to all images, mainly because of human-errors, and it is difficult to say that the image set is rotation or scale controlled (please see Figures \ref{operation} and \ref{dataset}). 

In addition to the product shooting and design file labelling efforts, an annotation effort was also carried out. For each photographed product the rectangle that encapsulates the product sample was annotated on the images (depicted as blue rectangles on Figure \ref{dataset}). These annotation will later be used as ground truth to our deep learning framework in the following section. 

In Figure \ref{dataset}, several examples from the image set are provided. As it can be seen from this figure, the set includes various types of \emph{background clutter, occlusions caused by packaging (tapes, chords, etc.), unwanted flash light reflections, non-uniform illumination, folding of the  sample product} and such. Thus, it is important that the matching solution we propose, must be robust to these types of effects. 

\section{Deep Product Segmentation}
\label{deepseg}
As mentioned in the introduction section, we apply a deep-segmentation supported bag-of-visual-words method to match product photos to design files. This framework, with rigorous benchmarking, is provided in the next section. However, before we get into the details of our matching framework, in this chapter we present some methods for segmenting the product region in product photos using  different deep CNN (DCNN) architectures. 

Finding the pixels that belong to a specific object category is known as \emph{semantic segmentation} in the literature. The reader may refer to various surveys on this problem \cite{Ahmad_2017,Jiangmedical2017,SiamEJY17,thoma2016,saffar2018semantic,YU201882,Guo2018,Garcia2017}. The literature involve hundreds of different approaches to semantic segmentation. The most common component among these approaches is undoubtedly the utilization of the abstract features of pre-trained DCNNs, by fine-tuning or transfer learning.

In this paper, in order to segment the pixels of a product photo using deep learning, we adapt three different architectures: ``FCN32s'', ``FCN8s'' and finally the proposed ``VGG-Regression-Net'', as we name it. 

FCN32s and FCN8s are well-known fully convolutional semantic segmentation networks, designed specifically for this problem \cite{Shelhamer2017}. They are originally trained for 21 different pixel labels. The only difference between these two architectures is that FCN8s includes skip connections that allow feature concatenation between different hierarchies within the DCNN. {In order to adapt these networks to our problem, the final deconvolutinal layers are set to 2 labels depth (as ``product'' and ``background''), and while this layer is being learned from scratch, all other convolutional layers in the networks are fine-tuned during training.}

\subsection{VGG-Regression-Net Architecture}

In addition to these fully convolutional architectures, a regression network is also proposed for the same problem. To that end, the convolution layers of a pre-trained DCNN, namely VGG-VD-19L \cite{Simonyan15} are transferred to a new structure, in which new fully connected layers with multiple neurons are added at the output layer and then trained. The aim is to assess whether a  DCNN with fully connected layers, trained for regression of the segmentation mask, performs better than fully convolutional architectures, such as FCN32s or FCN8s. We hypothesize that the fully connected layers can provide inference for a global composition of the image, in which the FCNs could fail to achieve. 

\begin{table}[t]
\centering
\begin{tabular}{|l|c|c|}
\hline
\multicolumn{1}{|l|}{\textbf{Input Layer}} & \multicolumn{2}{l|}{224$\times$224$\times$3 RGB Image. (VGG default input image size)}\\ \hline
\multicolumn{1}{|l|}{\textbf{VGG Layers}} &
\multicolumn{2}{|l|}{{Pre-trained VGG layers up to a K$^{th}$ layer}}\\ \hline \hline\hline
\multicolumn{1}{|l|}{\textbf{Appended Layers}} &
\multicolumn{2}{|l|}{}\\ \hline
\emph{layer no. - (type)}  & \emph{weight vector size} & \emph{output blob size}\\ \hline
\textbf{K+1} - (fully conn.)& m$\times$n$\times$f$\times$256 & 1$\times$1$\times$256\\ \hline
\textbf{K+2} - (fully conn.) & 1$\times$1$\times$256$\times$256 & 1$\times$1$\times$256\\ \hline
\textbf{K+3} - (fully conn.) &  1$\times$1$\times$256$\times$900 & 1$\times$1$\times$900\\ 
\hline
\multicolumn{1}{|l|}{\textbf{Output Layer}} & \multicolumn{2}{c|}{900x1 vector (30x30 Segmentation Result)}\\ \hline
\end{tabular}
\caption{The generic structure of the VGG-Regression-Net used in the experiments.}
\label{TheNet}
\end{table}

The detailed architecture of the VGG-Regression-Net is provided in Table \ref{TheNet}.  The initial pre-trained layers up to a selected layer of VGG-VD-19L are cut and the neuron outputs are transferred as inputs to new learning structures. The input layer of the original VGG-VD-19L receives 244$\times$224 pixels RGB images. So any product photo that is fed to this DCNN is first down-sampled into 244$\times$224 pixels resolution. 

In order to find the most suitable layer, multiple transfer learning experiments are run. During each separate experiment, a new CNN is created by transferring the VGG-VD-19L layers up to a selected layer and appending new fully connected decision layers, which create the segmentation mask for the product in the image. The learning rates for the transferred layers are set to zero, so their weights are kept constant during training. 

The ground truth of these masks are obtained by using the annotation mentioned in the previous section (please see the blue rectangles in Figure \ref{dataset}). For each product photo, a ground truth mask for which the pixels inside the rectangle region are 1 and the rest (i.e. the background) 0, is created and used for training. The output of the fully-connected network head consists of a vector with 900 components, representing 30$\times$30 pixels-sized segmentation mask. 

\subsection{Training the Deep Segmentation Architectures}
Although training the FCNs is a subject of semantic segmentation, training the proposed VGG-Regression-Net architecture corresponds to a multi-dimensional regression problem. In order to solve this problem {l$_1$-norm} is implemented as a loss function when training the DCNNs using back-propagation and stochastic gradient descent with momentum. Batch normalization is used with a batch size of 15 images\footnote{{Stochastic Gradient Descent (SGD) algorithm with momentum is employed, considering Momentum: 0.91, Initial Learning rate: 0.001 , Weight Decay: 0.0004.} MatConvNet \cite{MatConvNet} library is used for training {the VGG-Regression-Net, while MATLAB Deep Learning Toolbox is utilised for training the FCNs.}}.

For all architectures, data augmentation is applied by mirroring ($\times$2), zooming ($\times$2) and rotating ($\times$4) the product photos, thus enlarging the image set by 16. For training the architectures 75\% and for validation 15\% of the image set are used. For this reason, each experiment is executed 10 times, using a different subset for testing, which contains a separate 10\% of the whole image set.

\begin{figure*}[t]
\centering
\includegraphics*[clip=false,width=1\textwidth]{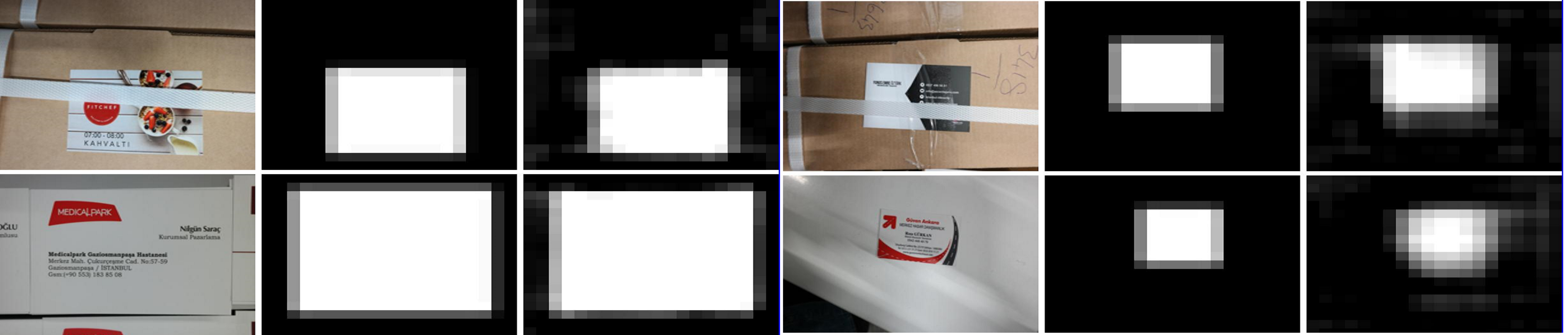}
\caption{Segmentation results for VGG-Regression-Net Layer 13, namely \emph{Conv5$^1$} are depicted. For four different samples, the image (left), the annotated ground truth (middle) and the DCNN output (right) are shown.}
 \label{deepsegresults}
\end{figure*} 

\begin{figure*}[t]
\centering
\includegraphics*[trim=12 451 382 265,clip=true,width=0.50\textwidth]{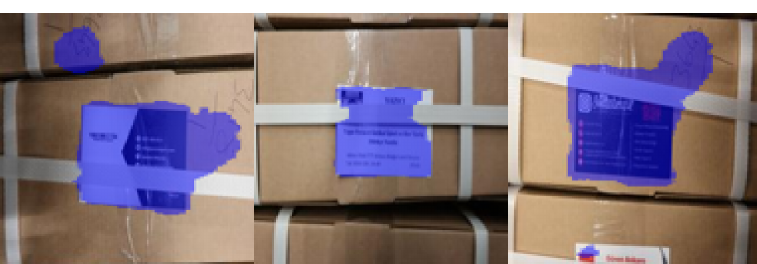}
\includegraphics*[trim=382 375 12 340,clip=true,width=0.47\textwidth]{Figure4.pdf}
\caption{Segmentation results, as blue regions over the image, for FCN8s (leftmost three images) and FCN32s (rightmost three images) are depicted. FCNs may create segmentation regions with disconnected blobs, since their fully convolutional nature has no means to prevent such an output. }
 \label{FCNs}
\end{figure*}

\begin{table}[t]
\centering
\begin{tabular}{|l|c|c|}
\hline
\textbf{Layer} & mean NCC & NCC standard deviation\\ \hline \hline
\textbf{VGG-Regression-Net Layer 09: Conv4$^1$} & 0.84 & $\pm$0.11\\ \hline
\textbf{VGG-Regression-Net Layer 10: Conv4$^2$} & 0.85 & $\pm$0.12\\ \hline
\textbf{VGG-Regression-Net Layer 11: Conv4$^3$} & 0.87 & $\pm$0.15\\ \hline
\textbf{VGG-Regression-Net Layer 12: Conv4$^4$} & 0.87 & $\pm$0.13\\ \hline
\textbf{VGG-Regression-Net Layer 13: Conv5$^1$} & 0.92 & $\pm$0.06\\ \hline
\textbf{VGG-Regression-Net Layer 14: Conv5$^2$} & 0.91 & $\pm$0.09 \\ \hline
\textbf{VGG-Regression-Net Layer 15: Conv5$^3$} & 0.90 & $\pm$0.14\\ \hline
\textbf{VGG-Regression-Net Layer 16: Conv5$^4$} & 0.89 & $\pm$0.12\\ \hline
\textbf{VGG-Regression-Net Layer 16: Conv5$^4$} & 0.89 & $\pm$0.12\\ \hline
\textbf{FCN32s} & 0.87 & $\pm$0.10\\ \hline
\textbf{FCN8s} & 0.85 & $\pm$0.09\\ \hline
\hline
\end{tabular}
\caption{For each experiment the mean and the standard deviation of the segmentation accuracy (i.e. normalized-cross correlation of test results with the ground truth) are calculated. The best performance is obtained for transfer learning at VGG-Regression-Net Layer 13.} 
\label{ConvNetresults}
\end{table}

Regarding the VGG-Regression-Net architecture, in order to find the layer that provides the best abstract features for segmentation, 8 different structures are trained, by cutting the VGG-VD-19L at 8 different layers, namely conv4$^1$, conv$4^2$, conv$4^3$, conv$4^4$, conv$5^1$, conv$5^2$, conv$5^3$ and conv$5^4$, which are the 9$^{th}$ to 16$^{th}$ layers of VGG-VD-19L. Thus for 8 layers of VGG-Regression-Net, FCN32s and FCN8s, separately for 10 test sets, a total of 100 learning experiments are run. 

In order to benchmark the success of different VGG-Regression-Nets and the FCNs architectures, normalized-cross correlation (NCC) of the test results with the ground truth are calculated and averaged over the entire set, respectively for each experiment. Table \ref{ConvNetresults} shows the average success rates for each test layer of VGG-Regression-Net and the FCNs. The best results are obtained using the 13$^{th}$ layer of VGG-VD-19L for fine-tuning experiments, with an average of 0.92 normalized cross-correlation. FCNs both perform poor. 

We believe that this is mainly because of the fact that, the problem we solve here is not exactly semantic segmentation. The product photo to be segmented, a poster, a flyer et cetera is a composition of objects, not a single object. Our problem is about learning the pixels of a product inside an image, as a composition of objects. We believe that the reason why the VGG-Regression-Net architecture performed better compared to FCNs is mainly because, fully connected layers can learn the global composition of abstract features, whereas FCNs, with limited receptive fields, search for objects in local regions. Therefore, as seen in Figure \ref{FCNs}, even disconnected blobs as segmentation results can be obtained for FCNs. 

In the rest of this paper, we examine the effect of these three different segmentation methods to our matching performance. For this purpose, each matching method is deep segmented by the two adapted FCNs and the best VGG-Regression-Net experiment, which is obtained by using the abstract features from layer 13 (namely conv$5^1$). 

\section{Image Matching Framework}
\label{ImgMatFWork}
As previously mentioned in the introductory sections, the aim of this study is to find a real-time solution to product photo and design file  matching problem. For this purpose we propose a framework with various benchmarking experiments. The proposed framework is hybrid in the sense that it fuses a conventional pattern recognition method that uses hand-crafted features with a deep segmentation technique. And while doing this, the study presents benchmarking of different methods, in order to correctly acknowledge the best performance for the given framework.  

\begin{figure}[t]
\centering
\includegraphics*[trim=10 20 10 12,clip=true,width=0.9\textwidth]{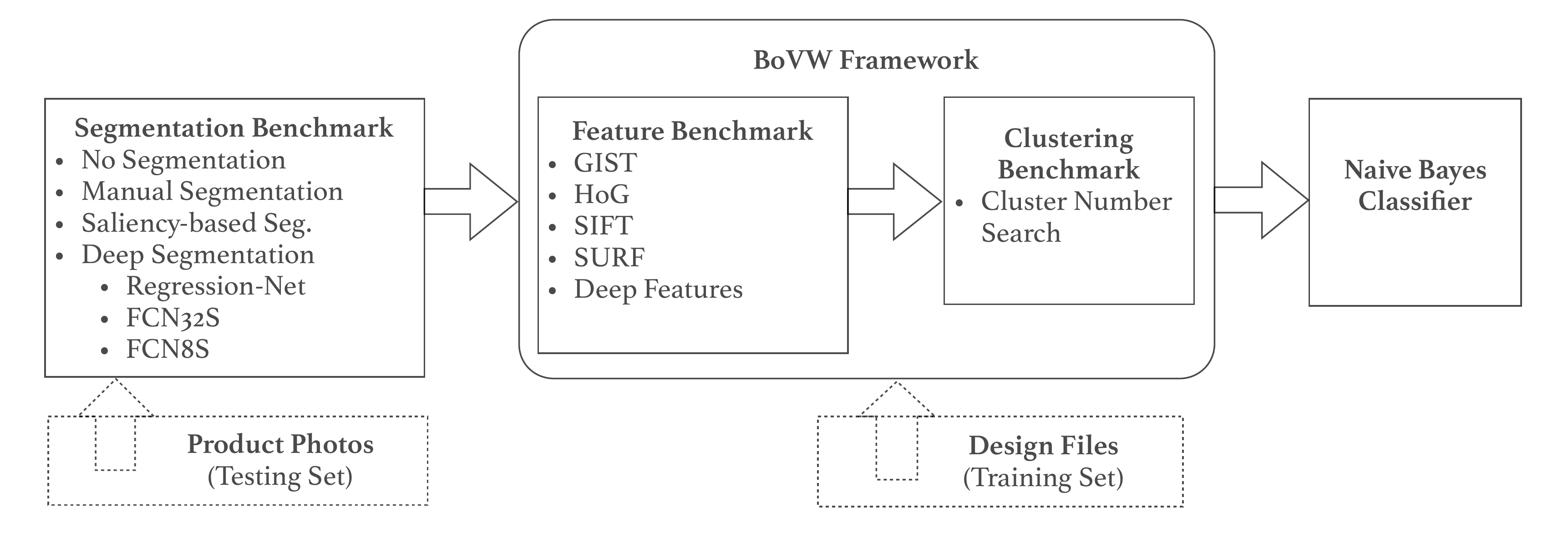}
\caption{The overall matching framework is depicted. The framework consists of four main blocks, namely segmentation, feature extraction, clustering and classification.}
 \label{fwork}
\end{figure}

In Figure \ref{fwork}, the high level depiction of our framework can bee seen with the benchmarking processes we utilize. The matching is accomplished by using the BoVW method \cite{Sivic2003} together with a Na\"ive Bayes classifier. The main advantage of using BoVW is that it provides a fixed-length representation of the image, regardless of the number or type of features obtained from that image. Moreover, BoVW + Na\"ive Bayes online operation (testing) is extremely fast. It requires a relatively slower offline training phase, in which the features obtained from training set is clustered into N sets. But needless to say, this does not affect the real-time operation in our framework. Within the BoVW framework, two principle benchmarking efforts are carried out, first being the benchmarking for selection of the hand-crafted or deep features of BoVW and second being the benchmarking for the optimal number of clusters for BoVW.

BoVW relies on the features obtained from a test image, which is a product photo in our case. The product photo does not only include the ``product'' but considerable background as well; ergo, it may be crucial to select the features only from the product region in the photo. For this purpose, as seen in Figure \ref{fwork}, another benchmarking effort for finding the product region is also carried out, using different segmentation methods including the deep segmentation techniques explained in the previous section. 

In the following subsections, we explain each benchmarking effort separately, following their process order in the framework. Thus, we start with the segmentation benchmark. Then we delve into our results and carry out discussions on the optimum method. 

\subsection{Segmentation Benchmark}
As mentioned above, selecting the features only from the product regions may dramatically affect the matching performance of a BoVW model. For this reason, in our experiments we utilize five product segmentation methods out of three categories and compare their results within the complete framework. We also include the case where no segmentation is carried out, so that we can clearly assess the contribution of a tested segmentation method.

\subsubsection{Manual Segmentation} We categorize our tested segmentation strategies in three titles, namely manual segmentation, unsupervised segmentation and supervised segmentation. Manual segmentation is accomplished by the operator. As seen in Figure \ref{operation}, the operator shoots the products with a mobile device and at this very moment, can manually segment the product region in the photo using the same mobile device. This is an unwanted scenario because it increases the operation time, which contradicts with the general purpose of the proposed system. However we still choose to utilize this segmentation method in our framework, so as to see the effect of ``\emph{perfectly}'' segmenting the product in a photo to our overall matching success and we regard this method as a ground truth for the segmentation step. In our image set we already have these manual annotations (please see Section \ref{imagesetsection} and Figure \ref{dataset}).

\subsubsection{Unsupervised Segmentation: Visual Saliency} The first automatic segmentation method we utilize is an unsupervised method to find the product region in a photo. Unsupervised segmentation had been a very hot topic \cite{ZHANG2008260} before  deep learning overwhelmingly manipulated the field with the idea of employing large-scale data to any problem. Since, for the sake of cheap and fast operation, we try to avoid deep operation as much as we can, we select a visual saliency-based method as our unsupervised segmentation method.

\begin{figure*}[t]
\centering
\includegraphics*[trim=0 0 0 0,clip=false,width=1\textwidth]{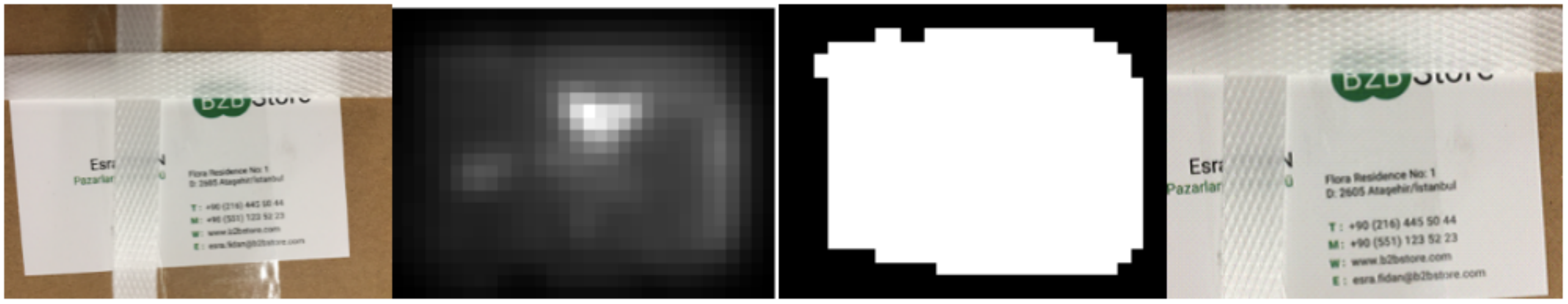}
\caption{The objects in the product photos are segmented using the GBVS algorithm \cite{Harel2006}, which is selected as the unsupervised segmentation method for our segmentation benchmark.}
 \label{gbvs}
\end{figure*}

For this purpose, we use Graph-based Visual Saliency (GBVS) method, which is a bottom-up visual saliency model. The algorithm creates Markov chains over the image map and treats the equilibrium distribution over map locations as saliency values \cite{Harel2006}. Although it is not a direct segmentation technique, visual saliency is being used as an objectness measure and is utilized for segmenting objects in an image \cite{7984578}. In Figure \ref{gbvs}, a sample result on a product photo can be seen. Firstly, the saliency heat map is calculated. Then by using a constant threshold (0.11 in our experiments), the object region is segmented. In the same figure, the segmented object can be seen in the rightmost image. 

There are various methods to segment an object from an image without any prior information, in other words in an unsupervised manner. The reason we choose to use GBVS is simply because of its speed-accuracy trade-off \cite{7984578}. In an extended study, it is possible to search for the most optimum method to segment an object in an unsupervised manner, however we find this effort beyond the scope of this study. 

\subsubsection{Supervised Segmentation: Deep Learning}
Supervision is simply utilizing domain-specific data. Thus, compared to any unsupervised method, it is more susceptible to over-fitting. However, if there is sufficient training data, supervised methods are preferable most of the time. In order to segment the product in a supervised manner, we utilize the three deep segmentation methods, explained in Section \ref{deepseg}. 
 
\subsection{Image Features Benchmark}
The general idea of BoVW is very simple: ``representing an image as a fixed-length set of features''. The so-called features consist of keypoints and descriptors. Keypoints denote the salient locations in the image, ideally invariant to transformations. Descriptor is the description ``\emph{around}'' the keypoint. BoVW use both keypoints and descriptors to construct vocabularies and represent each image as a frequency histogram of features that are in the image. Similarity measures to a test image can be calculated using these frequency histograms, and thus a classification can be performed. 

For a BoVW framework, the most important question is obviously ``which feature/descriptor to use''. Depending on the problem definition, imaging modality, performance requirements and computational budget, different methods can be used. For a comparison of local feature detectors and descriptors for visual object categorization, the reader may refer to \cite{Lankinen2012}. In this study we employ 5  popular features, namely, GIST  \cite{Oliva2001}, histogram of gradients (HoG) \cite{Dalal2005}, SIFT \cite{Lowe2004}, SURF \cite{Tuytelaars2006}, and deep features, which are obtained using a special CNN layer, namely the Spatial Pyramid Pooling (SPP) Layer \cite{He14SPP}. 

Among the aforementiond five features types, SIFT \cite{Lowe2004} and SURF \cite{Tuytelaars2006} are, by definition, \emph{local}; thus they are well-suited for BoVW. For HoG \cite{Dalal2005}, the locality should be pre-defined, i.e. provided by the user for local a region with a fixed area. We use blocks of 16x16 on a uniform grid and obtain HoG features individually from each block.

GIST \cite{Oliva2001}, on the other hand, is a global decriptor, more than a feature. It literally catches a ``\emph{gist}'' of the scene by using multi-scale low level features. Hence it is incompatible for a BoVW model, and accordingly it is implemented out of the BoVW framework. We calculate the GIST descriptors for each training and test data.  In consequence, by calculating the Euclidean distances between the GIST descriptors, we match a product photo to an image design file.  

Similarly to GIST descrpitor, the output of an SPP layer \cite{He14SPP} does not need a histogramization effort. This layer's output is already fixed-length. SPP collects activations from layers of different hierarchies, and concatenates them in a single fixed-length vector. {For this purpose, we have utilized combinations of activations from different layers of the FCN32s network as an input to the SPP layer}. By calculating the (weighted\footnote{In an SPP layer \cite{He14SPP}, the activations are weighted by the size of the pooling layer.}) Euclidean distances between these vectors, matching is performed. 

\subsection{Hyper-Parameter Optimization}
The proposed framework includes different methods with benchmarking of various intermediate steps (segmentation, feature extraction etc). Thus there are many hyper-parameters that may affect the system performance. In this study, we optimize only the cluster number of the BoVW framework. We believe that this is the most important hyper-parameter, mainly because it is independent of the utilized segmentation, feature extraction or the classification steps. The number of clusters is the vocabulary size and thus the heart of a BoVW model. Accordingly, in the next section, we also provide results for different cluster numbers, thus showing the effect of vocabulary size on performance.

\subsection{Classifier}
The final block of the proposed framework is classification. BoVW provides a fixed length histogram representation for any image, and classification within this vector space is another step, which serves as the final decision of the system. In our framework, the final goal is to find the design file that matches the given product photo. Various classification methods can be employed for a BoVW system and the reader may refer to \cite{Hentschel2014} for a detailed comparison. 

We have chosen Na\"ive Bayes algorithm as our classifier mainly because it is very fast and training requires a small
amount of samples to estimate the model parameters. That's why it has always been an optimal \cite{Kuncheva2006} partner for BoVW. We believe that with another, maybe mathematically more complex classifier, our results may further be improved. For the sake of computation speed we employ Na\"ive Bayes for all BoVW experiments in this study, and leave the benchmarking of different classifiers to a future study. 

\section{Experimental Results}
\label{ExpRes}
In this section, we present our experimental results. In two separate sections, we first provide the details of the experiment parameters and then present our comparative results with rigorous discussions.

\subsection{Experimental Setup}
The main objective of our experiments is to find an optimal method to product photo and design file matching. The absolute value of matching success depends on the number of design files to be compared in the image set. If there are only, for example, 10 design files to match, regardless of the benchmarked methods, the success would be relatively higher compared to a case in which thousands of possible design file candidates exist.

In our experiments, we have selected the number of product photos to match as 60\footnote{This parameter can be optimized with further experimentation. However this would require running thousands of experiments with 26 different methods, which we have chosen leave to a future study.}. Moreover, we have selected the number of design files as 100, so that the 60 product photos will match \emph{some} of the design files in this 100 element set, whereas the rest are just fillers\footnote{As explained in Section \ref{dataset}, the number design files that match a set of 60 product photos vary. A single page flyer has a single corresponding design file, whereas a two-sided business card has two design files matches for both sides. Thus the exact  number of fillers (i.e. randomly selected non-pair design files) in a set of 100 design files changes according to the set of product photos.}.

Accordingly, we have created 1000 different randomly selected 60 product photo - 100 design file sets, using the total 2000 product photos and 3458 design files. For each benchmarked method, 1000 experiments are run using these 1000 different test cases. 
At each experiment, for each product photo, the order of match is recorded. For this purpose, when a product photo is matched to 100 design files for an experiment, the Na\"ive Bayes probabilities (or the Euclidean distances) are calculated and sorted. The rank of the probability of the corresponding design file is recorded as the \emph{order} of that product photo's matching performance\footnote{For example, for a single product photo, we check the similarities to the given 100 design files in that experiment. The actual design file that matches the given product photo has the order 4, when all Na\"ive Bayes probabilities (or the Euclidean distances) are sorted. Then the \emph{order} for this product photo in this experiment is simply 4.}.

\subsection{Results}
For benchmarking, we employ 26 different methods using a combination of 6 different segmentation methods and 5 different features as presented in Section \ref{ImgMatFWork}. For instance, the case in which SIFT features, that are obtained from only the product region segmented using VGG-Regression-Net, is referred to as ``\emph{sift-deep}''. Or, the case, in which GIST global features are obtained from a manual segmented object region in a product photo, is called ``\emph{gist-manual}''. 

All 26 methods are tested for the same 1000 experiment sets and for each product photo in each experiment, the order of match is recorded. Using this order measure, we also calculate the average order of being matched. For example, for a specific method, the percentage of product photos that have smaller or equal order ``k'' is recorded as the \emph{{top-k accuracy}}. For example, if the \emph{{top-k accuracy}} for order k=10 is  95\%, this shows that by only checking best 10 possible match results, it is possible to match the correct design file with 0.95 probability.

\begin{figure*}[t]
\centering
\begin{subfigure}{8.1cm}	
	\includegraphics*[clip = true,trim=200 85 245 75,width=0.9\textwidth]{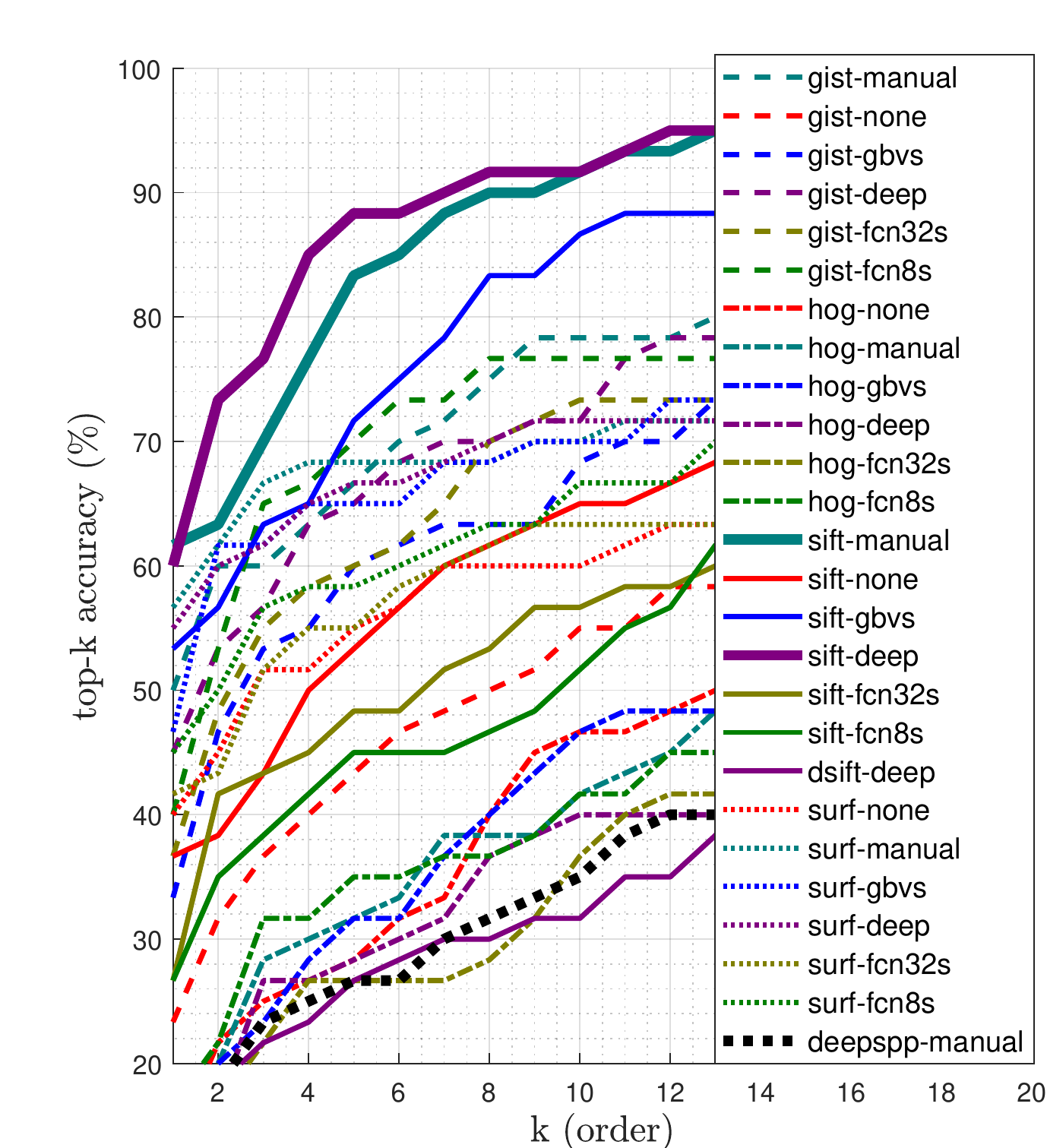}
\caption{\emph{Top-k accuracy} curves for all methods \\.}	
\end{subfigure}
\begin{subfigure}{8.1cm}	
	\includegraphics*[clip=false,trim=200 105 220 145,width=1\textwidth]{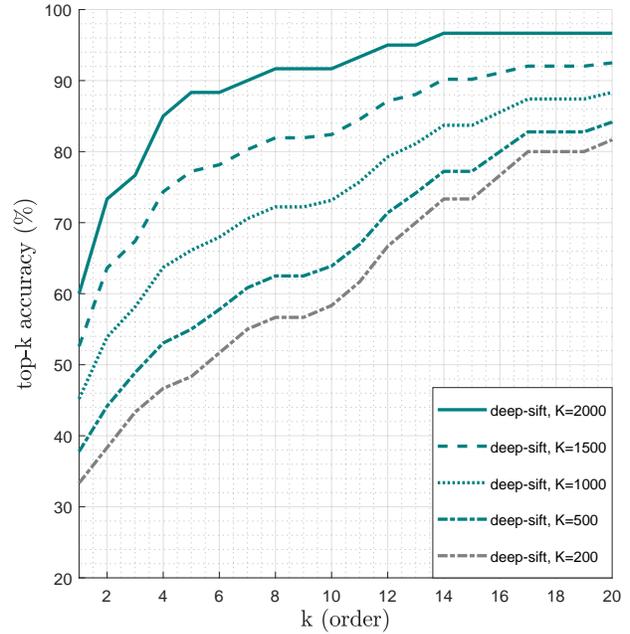}
\caption{\emph{Top-k accuracy} curves for \emph{sift-deep} under varying number of clusters in BoVW.}	
\end{subfigure}
\caption{Top-k accuracy curves are depicted. In these curve x-axis denotes the order of match, i.e. the order of similarity of the actual pain in the training set. The y-axis denotes the average success for that order value.}
 \label{Results}
\end{figure*}

In Figure \ref{Results}.a, the \emph{{top-k accuracy}} for all methods are depicted. The best accuracy is obtained with \emph{sift-deep} method, for which SIFT features that are obtained from only the product region segmented via the VGG-Regression-Net, are fed to the BoVW framework. The \emph{sift-deep} method also slightly outperforms \emph{sift-manual} method, for which the segmentation is performed by the human operator. This indicates that human operators can make mistakes in product region annotation, whereas deep learning-based segmentation method can generalize these errors and perform much better.

After observing the success of SIFT, we have applied another version of the SIFT descriptor, namely the ``\emph{Dense SIFT}'' (DSIFT) \cite{lenc2015understanding}. DSIFT is roughly equivalent to running SIFT on a dense grid of locations at a fixed scale and orientation. That's why DSIFT provide, on average, 10 times higher number of keypoints, compared to SIFT. In Figure \ref{Results}.a, the performance of DISFT, which is quite poor, can also be seen. This is, we believe, because of the fact that, increasing the number of keypoints does not support representation, but conversely creates more false alarm matches between feature clusters. 

The ``\emph{deepspp}'' method, in which deep CNN features are fed to a SPP layer, performs poor, even with manual (perfect) segmentation of the product. Deep features carry abstract information, which may fill the semantic gap of any vision problem. Consequently, this poor performance of deep features were intriguing for us. For this reason we have carried out extensive deep visualization experiments to uncover this issue. SPP creates activations from all (thousands even when only the deepest layer is used) neurons from the selected layers, most of which are unfortunately noise and are usually dropped out within the deep CNN. An SPP layer does not have the ability to select deep features according to their quality. That is why this method performs unsurprisingly poor {within a BoVW framework}, compared to a more selective and scale-invariant descriptor method, such as the SIFT.  We have utilized different combinations of activations from various layers of the FCN32s network. The best performance was obtained when only the activations from the final max-pooled convolutional layer (pool5 - 13$\times$13$\times$512) was utilized. Only the resulting curve for this case is depicted in Figure 7.a. Our visualisation experiments clearly show that, regardless of the layer the deep features are obtained, selective activations are always overwhelmingly outnumbered by noisy, unselective and insignificant activations, which can not lead to any semantic decision. 

Consistent with our observations presented in Section \ref{deepseg}, segmenting with FCN32s and FCN8s does not contribute the BoVW matching success positively. Semantic segmentation of product photos as if they are plain objects, is apparently not helping the feature selection operation enough. 

In Figure \ref{Results}.b, a parameter optimization effort for cluster numbers is depicted. As the number of clusters in BoVW increases, so as the success rates. In our tests, we tested up to 2000 clusters, which is the best case. This number can further be increased for higher success with a price of dramatically increasing our training time. The \emph{{top-k accuracy}} for  all methods that utilize BoVW in Figure \ref{Results}.a are calculated using 2000 clusters, which is our optimal case.

\begin{figure*}[t]
\centering
\includegraphics*[trim=156 260 156 236,clip=true,width=1\textwidth]{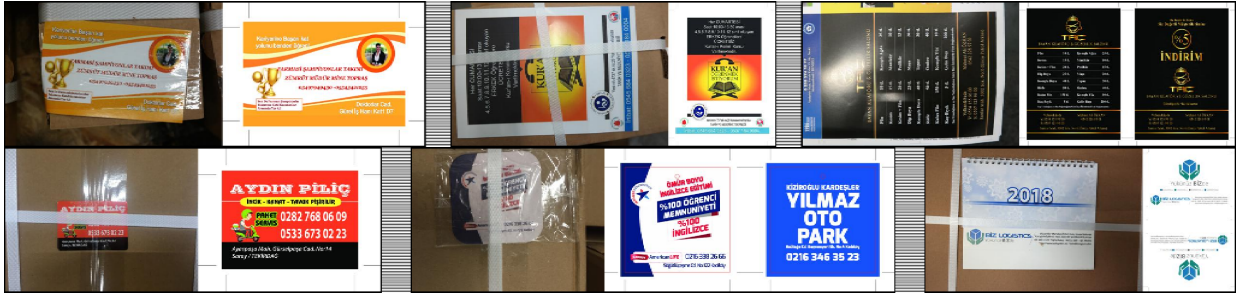}
\caption{Sample results for the \emph{sift-deep} method are seen. The samples on the top row are found with a perfect hit. The samples in the bottom row are matched with orders 8, 70 and 11 from left to right, respectively.}
 \label{sampres}
\end{figure*}

In Figure \ref{sampres}, some sample results for the \emph{sift-deep} method can be seen. The three samples on the top row are found with \emph{order} 1, i.e. with a perfect hit. The samples in the bottom row are matched with orders 8, 70 and 11 from left to right, respectively. The mismatch cases are usually because of strong clutter or impaired design files. 

The actual implementation of the system shows that the average ``end-to-end''  matching time for a product photo, using the \emph{sift-deep} method in a regular, no-GPU desktop computer is less than 4 seconds, including communication delays\footnote{(image upload: 0.65s) + (deep segmentation: 2.15s) + (SIFT extraction: 0.69s) + (matching 0.23s) + (downloading the results 0.025s) = (TOTAL 3.74s \emph{on average}). Feature extraction for design files are performed offline.}. This is a feasible duration for the operation considering that it is much faster than the operator manually searching for the product id, which takes about a minute for a single product photo. Still, the computation duration is open to improvement with better hardware and further software optimization.

\section{Conclusions}
In this paper, we propose a real-time image matching framework, which is hybrid in the sense that it uses both hand-crafted features and deep features obtained from a well-tuned deep convolutional network. We concentrate on a specific application, that is to say, printing design to product photo matching. Since photographs of a printed product suffer many unwanted effects, such as uncontrolled shooting angle, uncontrolled illumination, occlusions, printing deficiencies in color, camera noise, optic blur, et cetera, we benchmark different hand-crafted and deep features to choose an optimal performance and propose a framework, in which deep learning is utilized with highest contribution. 

Our results show that a deep segmentation supported BoVW method gives satisfactory results for the proposed operational concept. What is more, hand-crafted features, when deep segmented from a region of interest may lead to better results, compared to deep features, which may include overwhelming number of noisy and unselective activations.  

Like all current problems in computer vision, image matching problem is also moving to the DCNN domain. On the other hand, DCNNs require millions of data and expensive hardware. That's why we still need ingenious, practical and cheap industrial solutions until deep CNN hardware becomes standard in the following years.

In the meantime, we continue our studies on deep CNN structures, specifically on Siamese networks. We consider building a Siamese network which can learn similarities between a product photo and design file pair to be a novel study. Thus, we focus our studies on enlarging our image set for training of such a system. 
\section*{Acknowledgments}
This research was partially supported by the National Science Council of Turkey (TUBITAK - TEYDEB), with the project title {\emph{Customer-Information Matching, Invoicing and Barcoding of Post-Cut Products in Custom Printed Products using Image Processing Methods}}, under the 1507 program and with the project number {7170364}. The authors would like to thank the owner of the project, Şans Printing Industries (bidolubaski.com) for their support and hard-work.

\bibliographystyle{unsrt}
\bibliography{references}

\end{document}